\documentclass[lettersize,journal]{IEEEtran}
\usepackage{amsmath,amsfonts}
\usepackage{algorithmic}
\usepackage{algorithm}
\usepackage{array}
\usepackage[caption=false,font=normalsize,labelfont=sf,textfont=sf]{subfig}
\usepackage{textcomp}
\usepackage{stfloats}
\usepackage{url}
\usepackage{verbatim}
\usepackage{graphicx}
\usepackage{cite}
\usepackage{color}
\usepackage{amssymb}
\usepackage{booktabs}
\usepackage{makecell}
\usepackage{multirow}
\begin{document}

\title{A Survey on Datasets for Decision-making of Autonomous Vehicle}

\author{
\IEEEauthorblockN{ Yuning Wang, Zeyu Han, Yining Xing, Shaobing Xu*, Jianqiang Wang* }

\IEEEauthorblockA{School of Vehicle and Mobility, Tsinghua University, Beijing, China}
\thanks{Yuning Wang and Zeyu Han contribute equally to this work. This research was funded by the National Natural Science Foundation of China with award 52131201 (the key project) and Tsinghua University Toyota Joint Research Center for AI Technology of Automated Vehicle (TTRS 2023-06), the Intelligent driving and high-precision map key technology verification platform cooperation project.}}



\maketitle

\begin{abstract}
Autonomous vehicles (AV) are expected to reshape future transportation systems, and decision-making is one of the critical modules toward high-level automated driving. To overcome those complicated scenarios that rule-based methods could not cope with well, data-driven decision-making approaches have aroused more and more focus. The datasets to be used in developing data-driven methods dramatically influences the performance of decision-making, hence it is necessary to have a comprehensive insight into the existing datasets. From the aspects of collection sources, driving data can be divided into vehicle, environment, and driver related data. This study compares the state-of-the-art datasets of these three categories and summarizes their features including sensors used, annotation, and driving scenarios. Based on the characteristics of the datasets, this survey also concludes the potential applications of datasets on various aspects of AV decision-making, assisting researchers to find appropriate ones to support their own research. The future trends of AV dataset development are summarized.
\end{abstract}


\section{Introduction} \label{Sec:Introduction}
\IEEEPARstart{T}{he} technology of \textcolor{black}{Autonomous Vehicles (AV)} is placed great expectations of rising revolution in people’s travel manner and future transportation system \cite{wang2021towards}. The architecture of \textcolor{black}{autonomous driving} technology can be roughly divided into three main modules: environmental perception, decision making, and motion control  \cite{schwarting2018planning} \cite{paden2016survey}. In particular, the decision-making module plays the role of the driver’s brain with great significance.

Along with the rapid development of \textcolor{black}{AV} technology, the research on decision-making is also on the rise \cite{paden2016survey} \cite{duan2020hierarchical} \cite{badue2021self}. Decision-making is usually decomposed into hierarchical modules, including global route planning, behavior selection, and trajectory generation \cite{badue2021self}. Their boundaries were explicit but are becoming obscure as the deep learning and end-to-end approaches emerge \cite{schwarting2018planning}\cite{wang2023decision}. In this paper, we use a relatively generalized definition of decision-making, which includes not only motion planning but also environment and driver understanding.

\begin{figure}[!t]
\centering
\includegraphics[width=\linewidth]{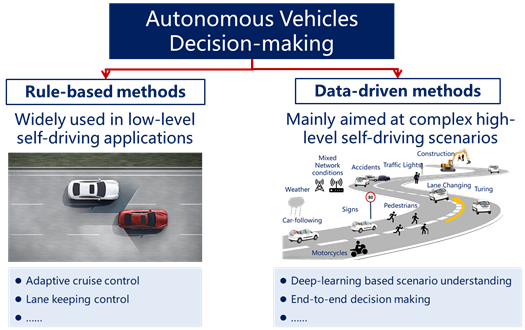}
\caption{Two types of autonomous vehicle decision-making algorithms}
\label{Fig:decision level}
\end{figure}

The technique of decision-making can be driven by physics/optimization-based rules, or by machine intelligence learned from data, as shown in Fig. \ref{Fig:decision level}. As for low-level \textcolor{black}{autonomous driving, rule-based approaches} are more widely used to achieve \textcolor{black}{driving} capacity in simple scenarios. However, toward high-level \textcolor{black}{autonomous driving} technology, rule-based approaches are usually considered to lack generalizability in complex traffic situations \cite{sadat2019jointly}. To this end, data-driven approaches, including machine learning, deep learning, and reinforcement learning, have aroused researchers’ interest. The quantitative goals of these approaches can be classified into two categories. \textcolor{black}{Some studies treated human driving data as expert knowledge and ground truth so that the task is to minimize the error to the real driving \cite{krajewski2018highd}\cite{sun2020scalability}. Some other studies used mixed goal functions consisting of errors to real trajectories and objective metrics such as driving risk and time efficiency \cite{bronstein2022hierarchical}\cite{wang2023differentiated}.} Data-driven approaches outperform the rule-based approaches in some complex scenarios, as reported in \cite{liang2018cirl}, but their performance is actually restricted by the datasets used to train and test the algorithms. To guarantee unbiasedness and generalization of these approaches, proper and abundant data is essential and dominants data-driven decision making.

\begin{figure}[!t]
\centering
\includegraphics[width=\linewidth]{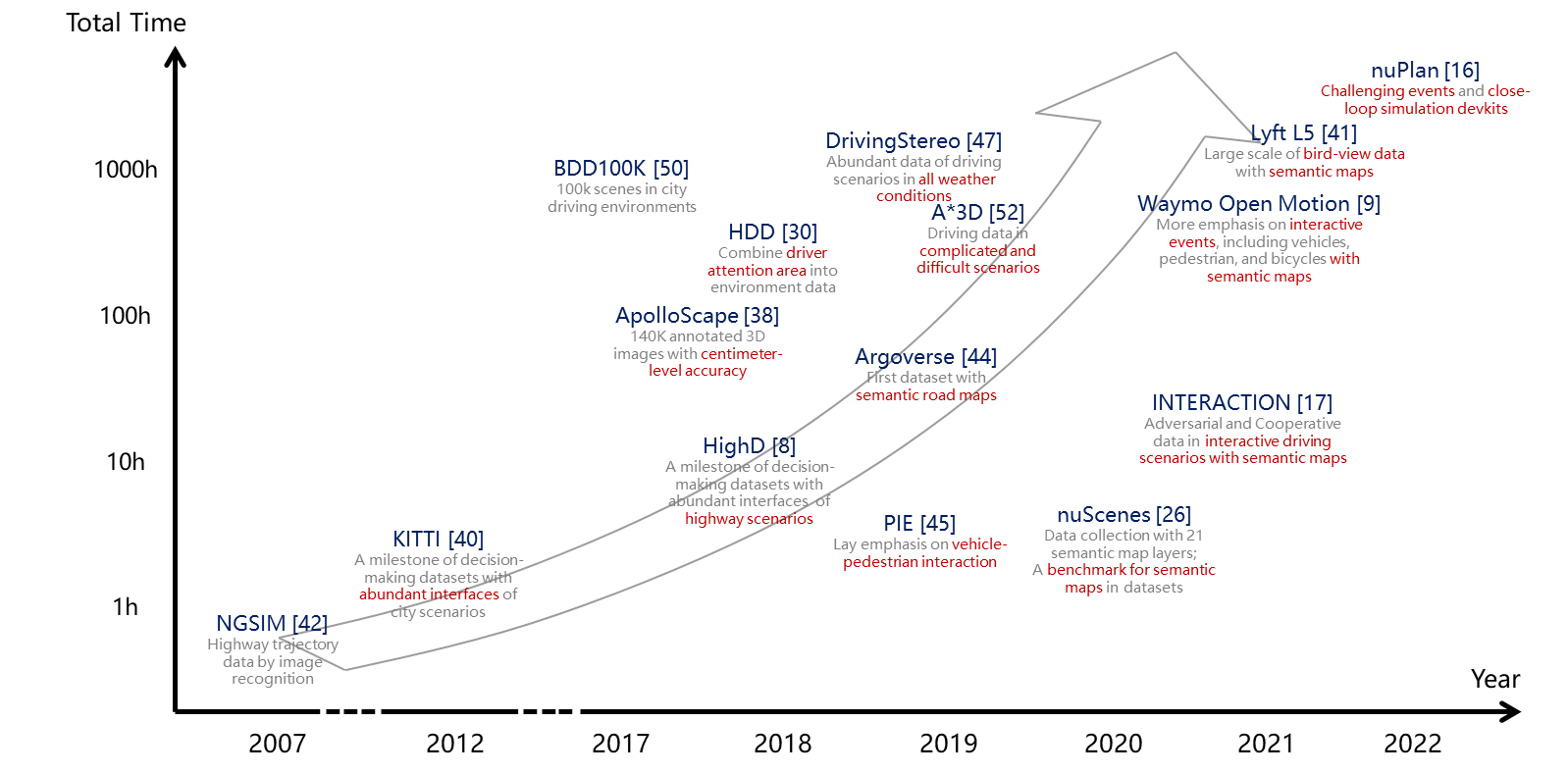}
\caption{\textcolor{black}{The size trend of decision-making datasets for autonomous vehicle}}
\label{Fig:dataset size}
\end{figure}

\begin{figure*}[!t]
\centering
\includegraphics[width=\linewidth]{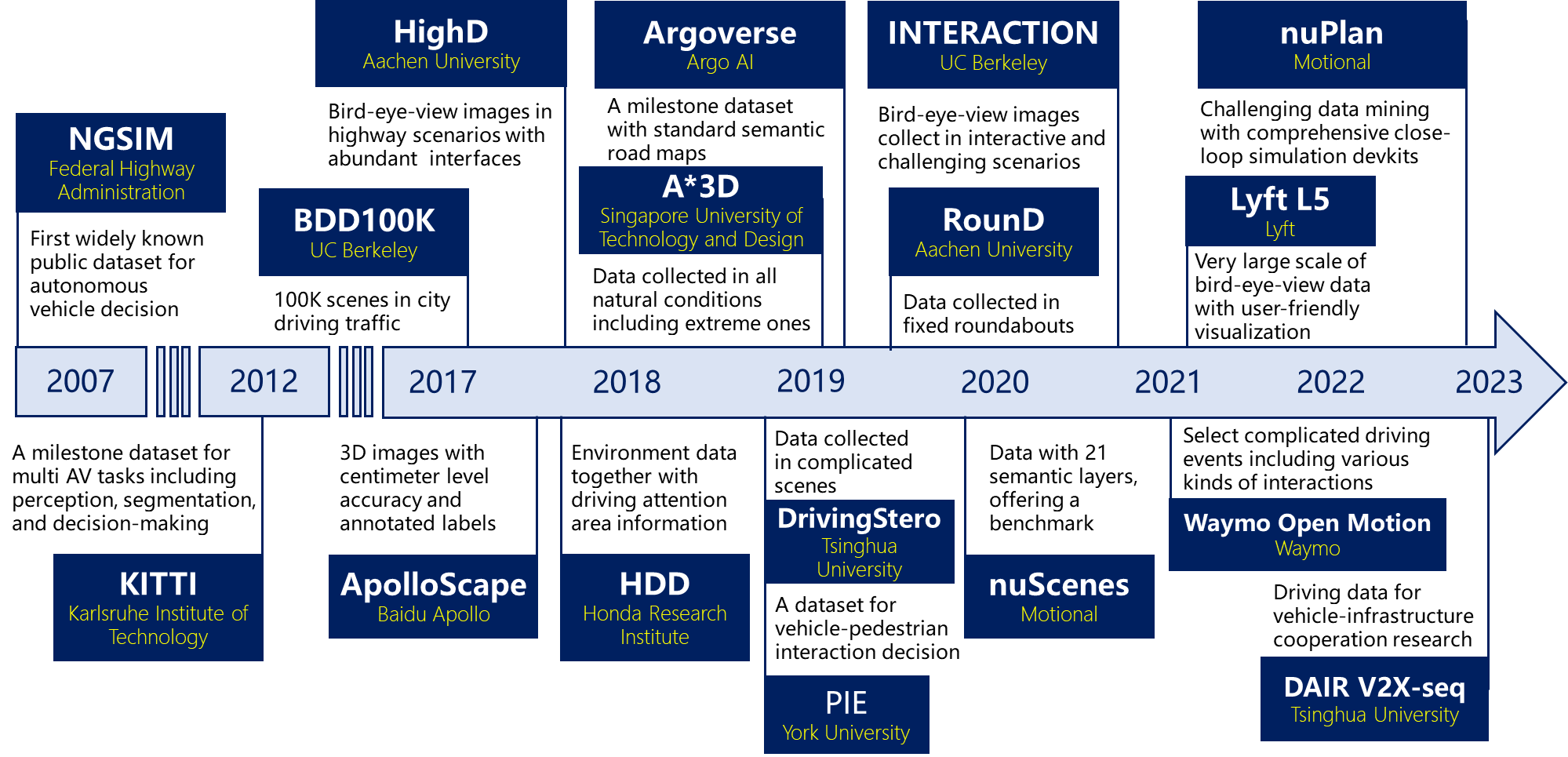}
\caption{\textcolor{black}{The developing history of autonomous vehicle decision-making datasets}}
\label{Fig:dataset history}
\end{figure*}

A number of large-scale datasets for decision-making have be created recently. \textcolor{black}{The common features of them are that basic final decision outputs, which can be expressed by trajectories, control quantities, and driving behavior labels, are included. However, the data sizes, formats, categories, and applications vary greatly.} Fig. \ref{Fig:dataset size} demonstrates the size explosion of decision-making datasets and the rapid growth of demands. From the aspect of data volumes, it exploded by orders of magnitude in the last 15 years, growing from one hour to over 1000 hours with much richer scenarios. Fig. \ref{Fig:dataset history} shows a detailed developing trend of decision-making datasets, including name, developer, and main features of each dataset. Some of them have become the  milestones in this field. The data types and characteristics related to decision-making are various. Some focus on vehicle status \cite{geyer2020a2d2}, some concern about environmental understanding \cite{patil2019h3d}, while some pay attention to drivers’ behavior analysis \cite{alletto2016dr}. The recent datasets have laied more emphasis on challenging scenarios instead of regular ones \cite{Caesar2021nuplan} \cite{sun2020scalability}\cite{zhan2019interaction} \cite{Krajewski2020rounD}. Another obvious trend is that recently emerging datasets had much larger data volume, and the types and sources of data are also more varied. Therefore, surveying these latest datasets matches the increasing demand.

Recently, Yin Hang and Christian Berger presented an overview of 27 existing publicly available \textcolor{black}{autonomous driving} datasets and provided comparisons and guidelines for researchers to select datasets \cite{yin2017use}. The methodologies of collecting datasets and the highlights of each dataset are also listed in that paper. However, the information and analysis provided are relatively less complete. As an extension, \cite{kang2019test} analyzed virtual testing environments in addition to real on road or in-track testing. Junyao Guo et al. categorized the datasets and highlighted the datasets with complicated environments \cite{guo2019safe}. The article also introduced the factors and metrics of drivability to measure the adaptability of autonomous driving in various driving conditions. Di Feng et al. demonstrated datasets for deep multi-modal object detection and semantic segmentation \cite{feng2020deep}. As a summary, most \textcolor{black}{autonomous driving} datasets and correlative surveys concentrated on environmental perception technology such as object detection and object tracking \cite{kenk2020dawn} \cite{pang2020tju}. However, attention to datasets for decision making is insufficient. Therefore, a comprehensive survey focusing on datasets for decision-making is needed.

\textcolor{black}{In this paper, we survey the datasets related to AV decision-making development. Apart from the basic common decision output information, we classify the datasets into three categories based on the sources of influencing factor data recorded in them, which are vehicle data, driving environment data, and driver data. The potential applications are also concluded after the introductions to the datasets.} Compared with previous surveys of \textcolor{black}{autonomous driving} datasets, the major contributions of this paper are as follows: 

\begin{enumerate}
    \item This paper presents a detailed summary of \textcolor{black}{AV decision-making datasets}, especially, many latest datasets that are not covered by other previous papers are included, which benefits researchers in this area to obtain a more focused view of related datasets.
    \item The datasets are categorized by the data sources, i.e., drivers, vehicles, and environment. This categorization provides a criterion for classifying and comparing different datasets. Besides, various data types and sources classified in this categorization are convenient for researchers to choose datasets adaptive to their sensors and research demand.
    \item This paper summarizes the application purposes of datasets for decision-making. This information helps researchers to choose appropriate datasets for different research.
\end{enumerate}

The remainder of this paper is organized as follows. Section \ref{Sec:Vehicle Datasets}-\ref{Sec:Driver Datasets} introduces the decision-making datasets in the order of vehicle-related datasets, environment-related datasets, and driver-related datasets, respectively. Section \ref{Sec:Dataset Applications} demonstrates dataset application, and Section \ref{Sec:Future Trends} discusses the future trend of decision-making datasets. The conclusion is provided in Section \ref{Sec:Conclusion}.

\section{Vehicle Datasets} \label{Sec:Vehicle Datasets}
Data from vehicles \textcolor{black}{reflects the basic status and motions of vehicles}. Therefore, the related datasets have wide applications in research on \textcolor{black}{AV} decision-making. The collection of vehicle-related data is more convenient than environment or drivers related data, but the analysis is relatively complex. In this section, 11 datasets including data types and collection platforms are reviewed.

\subsection{Datasets Classification}
\begin{figure}[!t]
\centering
\includegraphics[width=\linewidth]{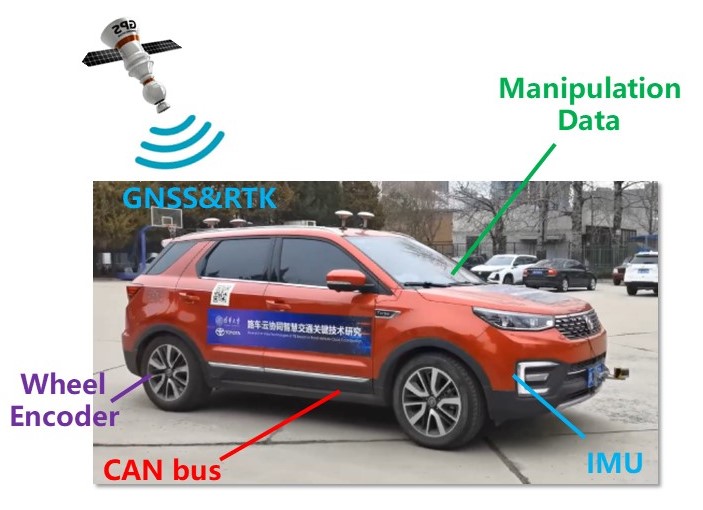}
\caption{Typical vehicle data}
\label{Fig:vehicle data}
\end{figure}

There are mainly three classes of vehicle-related data, as shown in Fig. \ref{Fig:vehicle data}.

\begin{enumerate}
    \item Status data: \textcolor{black}{reflecting} the location, speed, and other external status information of the vehicle, such as distance, speed, yaw rate, and acceleration from \textcolor{black}{GPS (Global Positioning System) / GNSS (Global Navigation Satellite System), RTK (Real Time Kinematic), IMU (Inertial Measurement Unit)} and wheel speed sensors.
    \item Manipulation data: \textcolor{black}{containing} drivers' manipulation information, including steering angle, steering rate, accelerator pedal position, and brake pedal position data.
    \item Internal data: \textcolor{black}{recording} the internal status of the vehicle, such as the engine speed, the liquid pressure of brake valves, the transmission gear position, and other valuable data on the \textcolor{black}{CAN (Controller Area Network) bus}. 
\end{enumerate}

The status data is usually collected along with the perception data such as LiDAR and camera data, which acts as the input data of the decision-making algorithm. The manipulation data reflects the final results of decision-making and usually serves as the ground truth of the decision-making algorithm. Besides, the internal data contains underlying information, which sometimes is also used as the input of  learning-based algorithm. In Table \ref{Table:Vehicle Datasets}, 11 relative \textcolor{black}{autonomous driving} datasets for decision-making are listed as a comparison.

\begin{table}[!htbp]
    \centering
    \caption{Data Classes in Vehicle Datasets}
    \resizebox{\linewidth}{!}{
    \begin{tabular}{cccc} 
        \toprule
        Dataset \qquad &Status Data \qquad &Manipulation Data \qquad &Internal Data \\
        \midrule
        \makecell[c]{EU Long-term \cite{yan2020eu}}  &\checkmark &$\times$ &$\times$ \\
        \textcolor{black}{nuPlan \cite{Caesar2021nuplan}} &\textcolor{black}{\checkmark} &\textcolor{black}{$\times$} &\textcolor{black}{$\times$}\\
        nuScenes \cite{caesar2020nuscenes} &\checkmark &$\times$ &$\times$ \\
        KAIST \cite{choi2018kaist} &\checkmark &$\times$ &$\times$ \\
        BDDV \cite{xu2017end}  &\checkmark &$\times$ &$\times$ \\
        Udacity \cite{Udacity} &\checkmark &\checkmark &$\times$ \\
        A2D2 \cite{geyer2020a2d2} &\checkmark &\checkmark &$\times$ \\
        HDD \cite{ramanishka2018toward}  &\checkmark &\checkmark &$\times$ \\
        DBNet \cite{chen2019dbnet} &\checkmark &\checkmark &$\times$ \\
        DDD17 \cite{binas2017ddd17} &\checkmark &\checkmark &\checkmark \\
        DDD20 \cite{hu2020ddd20} &\checkmark &\checkmark &\checkmark \\
        D$^2$CAV \cite{toghi2020maneuver} &\checkmark &\checkmark &\checkmark \\
        \bottomrule
    \end{tabular}
    \label{Table:Vehicle Datasets}}
\end{table}

A group of datasets includes status data only. Their location data (GPS/IMU) and odometry data are mostly combined with other sensors such as cameras and LiDAR to design perception and localization algorithms. The EU Long-term dataset \cite{yan2020eu} collects GPS and IMU data along with LiDAR, camera, and radar data, allowing the vehicle to perceive its surroundings and locate itself. The data is collected by a multisensory platform and \textcolor{black}{ROS (Robot Operating System)} as the middleware. Similarly, the KAIST dataset \cite{choi2018kaist} records GPS, IMU, and wheel encoder data for \textcolor{black}{SLAM (Simultaneous Localization and Mapping)} algorithm design. Besides, BDDV \cite{xu2017end} is a dataset comprised of real driving videos, vehicle course and speed, and data from GPS, IMU, gyroscope, and magnetometer. These modalities can be used to recover the trajectory and dynamics of the vehicle to serve as the ground truth of end-to-end \textcolor{black}{autonomous driving} algorithms.

However, to learn an end-to-end \textcolor{black}{autonomous driving} algorithm, it is more appropriate to learn from vehicle manipulation instead of the trajectory only. A group of datasets involves the manipulation data for this purpose. The Udacity dataset \cite{Udacity} collects camera, LiDAR, GPS, IMU, and the manipulation data from CAN bus including steering angle and brake/throttle pedal opening, which can be used to train a steering control network or an image-based localization algorithm. The A2D2 dataset \cite{geyer2020a2d2} also includes manipulation data as labels for reinforcement learning research. Besides the end-to-end algorithms, numerous research pays attention to behavioral decision learning, and several datasets emerge consequently. The HDD dataset \cite{ramanishka2018toward} records CAN data including throttle opening, brake pressure, steering angle, yaw rate, and speed, together with location information and data from other sensors. The DBNet dataset \cite{chen2019dbnet} is also a large-scale dataset containing driver manipulation in a variety of traffic conditions for driving behavior learning.

Since manipulation behavior is usually correlated with of the internal status of vehicles, such as accelerating corresponds to increasing engine speed, a number of datasets collect not only the vehicle’s kinematic status and drivers’ manipulation data, but also the vehicle’s internal data. The DDD17 dataset \cite{binas2017ddd17} records abundant types of internal data including the engine speed, torque input to the transmission, fuel consumption, and gear position of transmission along with other kinematic status data and manipulation data. These data including the images from an event camera enables researchers to perform end-to-end learning of steering control. Based on DDD17, the DDD20 dataset \cite{hu2020ddd20} expands the data duration from 12h to 39h. Note that these two datasets both do not contain LiDAR, radar, and other sensors necessary for a complete \textcolor{black}{autonomous driving} solution. Toghi et al. \cite{toghi2020maneuver} discover the fact that the driver behavior distribution in most datasets is imbalanced, thus presenting the D$^2$CAV dataset from human-driven vehicles performing evenly distributed behaviors, containing CAN data such as engine speed for manipulation prediction purposes.

\subsection{Limitations}
Taking an overview of datasets for decision-making of \textcolor{black}{AV}, the limitations of existing datasets can be summarized into two main aspects: the lack of driver data and the lack of standards of data type.

\subsubsection{The lack of driver data}
Most datasets containing vehicle manipulation and internal status data are created mainly for end-to-end learning research such as predicting the steering angle based on the camera or LiDAR data. However, these algorithms have a non-negligible disadvantage of interpretability. Recently, plenty of researchers have paid their attention to driver’s behavior learning \cite{schwarting2018planning}, which learns behaviors such as overtaking, car-following, and lane-changing from human driving data. These behaviors can be easily understood by human drivers, and can be even considered as a priori knowledge for learning the motion or manipulation of the vehicle. This promising research area calls for datasets containing not only data from perception sensors and CAN bus, but also from the drivers, including the driver behaviors, postures, and attention areas. D$^2$CAV \cite{toghi2020maneuver} is a comparatively comprehensive dataset containing driver behavior, but the scenarios are relatively limited. As for driver postures or attentions, they are mostly studied for driver status recognition, and there has been no dataset yet that collects them along with perception sensor data and vehicle data for driver behavior learning research.

\subsubsection{The lack of data type standards}
Data from vehicles are of great variety, especially the CAN bus data. It is often simple to collect  all relevant data from the CAN bus, but specifying which types of data are more useful for \textcolor{black}{autonomous driving} decision-making and building comprehensive data standards for datasets is complex yet important work. The lack of data type standards also limits the comparison between different decision-making datasets. The construction of data type standards for datasets is a holistic study that is noteworthy for the whole community of autonomous vehicle research.

\section{Driving Environment Datasets} \label{Sec:Driving Environment Datasets}
Datasets related to driving environment have been studied extensively since the 21st century to accelerate vehicle technology development \cite{guo2019safe}. With the breakthrough of hardware and neural networks, a large number of complicated datasets have been proposed with varieties of sensors, data formats, and scenarios \cite{kang2019test} \cite{liu2021survey}. The features are so diverse that it is confusing for developers to choose the most appropriate dataset for a specific study where data mining is a must for research.  Consequently,  this section reviews those environment-related datasets and highlights their features.

\subsection{Data Types and Annotations}
While using datasets for academic or engineering purposes, data types and annotation labels are two of the features that users are most concerned about, especially for driving environment-related tasks since the real driving scenarios are so complicated and diverse. Only if the input and output interfaces are included in the datasets could the model be trained. In Table \ref{Table:Environment Datasets Sensors} and \ref{Table:Environment Datasets Annotations}, overviews of sensors used and annotation labels are given, illustrating the features of existing datasets. In Table \ref{Table:Environment Datasets Sensors}, Cam is the abbreviation of Cameras.

\begin{table}[!htbp]
    \centering
    \caption{Sensors on Collection Platforms of Driving Environment Datasets}
    \resizebox{\linewidth}{!}{
    \begin{tabular}{cccccc} 
        \toprule
        Dataset &Cam &Radar &LiDAR &\textcolor{black}{Roadside Sensors} &Others\\
        \midrule
        \textcolor{black}{MONA \cite{gressenbuch2022mona}} &\textcolor{black}{\checkmark} &\textcolor{black}{$\times$} &\textcolor{black}{$\times$} &\textcolor{black}{$\times$} & \\
        \textcolor{black}{nuPlan \cite{Caesar2021nuplan}} &\textcolor{black}{\checkmark} &\textcolor{black}{$\times$} &\textcolor{black}{\checkmark} &\textcolor{black}{$\times$} & \\
        \textcolor{black}{DAIR V2X-seq \cite{yu2023v2x}} &\textcolor{black}{\checkmark} &\textcolor{black}{$\times$} &\textcolor{black}{\checkmark} &\textcolor{black}{\checkmark} & \\ 
        ApolloScape \cite{huang2018apolloscape} &\checkmark &$\times$ &\checkmark &$\times$ & \\
        nuScenes \cite{caesar2020nuscenes} &\checkmark &\checkmark &\checkmark &$\times$ & \\
        WAYMO \cite{sun2020scalability} &\checkmark &$\times$ &\checkmark &\checkmark & \\
        WoodScape \cite{yogamani2019woodscape} &\checkmark &$\times$ &\checkmark &$\times$ & \\
        KITTI \cite{geiger2012we} &\checkmark &$\times$ &\checkmark &$\times$ & \\
        Lyft L5 \cite{houston2020one} &\checkmark &$\times$ &\checkmark &\checkmark & \\
        NGSIM \cite{kovvali2007video} &$\times$ &$\times$ &$\times$ &\checkmark & \\
        HighD \cite{krajewski2018highd} &$\times$ &$\times$ &$\times$ &\checkmark & \\
        ROAD \cite{singh2021road} &\checkmark &$\times$ &$\times$ &$\times$ & \\
        Argoverse \cite{chang2019argoverse} &\checkmark &$\times$ &\checkmark &$\times$ &Stereo Cameras \\
        PIE \cite{rasouli2019pie} &\checkmark &$\times$ &$\times$ &$\times$ & \\
        TITAN \cite{malla2020titan} &\checkmark &$\times$ &$\times$ &$\times$ & \\
        DrivingStereo \cite{yang2019drivingstereo} &\checkmark &$\times$ &\checkmark &$\times$ & \\
        ONCE \cite{mao2021one} &\checkmark &$\times$ &\checkmark &$\times$ & \\
        H3D \cite{patil2019h3d} &\checkmark &$\times$ &\checkmark &$\times$ & \\
        HDD \cite{ramanishka2018toward} &\checkmark &$\times$ &\checkmark &$\times$ & \\
        CADC \cite{pitropov2021canadian} &\checkmark &$\times$ &\checkmark &$\times$ &Thermal Cameras \\
        BDD100K \cite{yu2020bdd100k} &\checkmark &$\times$ &$\times$ &$\times$ & \\
        INTERACTION \cite{zhan2019interaction} &$\times$ &$\times$ &$\times$ &\checkmark & \\
        PREVENTION \cite{izquierdo2019prevention} &\checkmark &\checkmark &\checkmark &$\times$ & \\
        DDD20 \cite{hu2020ddd20} &\checkmark &$\times$ &$\times$ &$\times$ &Event Cameras \\
        A*3D \cite{pham20203d} &\checkmark &$\times$ &\checkmark &$\times$ & \\
        \bottomrule
    \end{tabular}}
    \label{Table:Environment Datasets Sensors}
\end{table}

As for sensors, the most chosen combination is multiple cameras and a multi-thread LiDAR. Together with images and point clouds, the driving environment can be reproduced precisely \cite{patil2019h3d}. Cameras ensure the semantic target classification and LiDAR is usually responsible for depth matching \cite{yang2019drivingstereo}. Some datasets mainly aim at pattern analysis tasks where positioning is not a must, so merely using multiple cameras without LiDAR is also a common configuration  \cite{rasouli2019pie} \cite{malla2020titan}. With the development of connected transportation, it is easier to establish a precise mapping of the real traffic environment  with the help of emerging perception technologies. Krajewski et al \cite{krajewski2018highd} concluded four methods for data capture, e.g., using drones and infrastructure-based sensors. This method can capture the whole scenario and is a good complement of onboard sensors. The datasets including Lyft L5 \cite{houston2020one}, INTERACTION \cite{zhan2019interaction}, Top-view Trajectories \cite{hu2020ddd20}, etc., took advantage of roadside units and drones to generate precise global virtual environments.  \textcolor{black}{DAIR V2X-seq \cite{yu2023v2x} and MONA \cite{gressenbuch2022mona} also used fixed sensors including LiDARs and Cameras on infrastructures to collect vehicle data from roadside view, which could be an additional data source for group decisions of connected agents.} In addition, to adapt to specific scenarios, some specialized sensors are utilized. Argoverse \cite{chang2019argoverse} used stereo cameras to form spatial features. CADC \cite{pitropov2021canadian} includes thermal cameras to perceive objects in snowy environments. DDD20 \cite{hu2020ddd20} proposed that event cameras contribute hugely to capturing dynamic environment variations. In conclusion, diverse sensor combinations determine the various features and intentions of datasets. 

Apart from the raw data, the annotation label is another important factor users consider for model training and development. We conclude four classes of annotations in environment-related datasets for decision-making research, which are bounding boxes in first-person view (FPV), physical states in bird eye view, high-definition (HD) map, and behavior labels \cite{guo2019safe}. Bounding boxes in FPV are abundant in almost all recent AV datasets due to the breakthrough in the field of computer vision. Many auto-label models have been developed to offer precise label information \cite{patil2019h3d} \cite{singh2021road} . Compared with the FPV bounding boxes, the physical states in bird-view (BV) are more helpful for developing decision methods of high-level \textcolor{black}{autonomous driving}, since targets in the blind spots of self-vehicle’s vision need to be considered \cite{wang2021reinforcement}. Some typical datasets including INTERACTION \cite{zhan2019interaction} and Lyft L5 \cite{houston2020one} generate bird-view information by roadside sensors or drones directly. And other datasets like WAYMO OPEN MOTION \cite{sun2020scalability}, H3D \cite{patil2019h3d} combined various resources to create a comprehensive digital twin of the whole driving scenarios. HD map refers to maps with lane-level recognition including drivable areas, obstacles, etc. \textcolor{black}{nuScenes \cite{caesar2020nuscenes} input the semantic map of collected scenes for further application with the dataset. On the basis of nuScenes, nuPlan \cite{Caesar2021nuplan} established a closed loop simulation system, allowing interactive decision algorithm validation.} DrivingStereo \cite{yang2019drivingstereo} used stereo cameras to derive a colorized disparity map containing semantic information. ONCE \cite{mao2021one} generated the HD map with the SLAM technology.

Behavior labels are one of the focuses in recent years since more and more deep-learning decision-making models need them for ground truth or intermediate variables. For instance, Toghi et al \cite{toghi2020maneuver} and Jain et al \cite{jain2016brain4cars} label events into discrete behaviors including turns, lane shifts, deceleration, etc. However, abstracting the whole events into one single label greatly compresses the amount of information, which is less accurate actually. Comparatively, HDD invented a 4-layer annotation structure, decomposing an event into target goals, behaviors, reasons, and attention area, and each layer corresponded to a finite label set \cite{ramanishka2018toward}. Similarly, Singh et al \cite{singh2021road} summarized an event with a triplet structure consisting of active agents, actions it performs, and the corresponding event locations in the ROAD dataset. Compared with one single label, sufficient antecedents and consequences express a traffic event better, making it easier for the training model to achieve precise decisions and possess interpretability.

\begin{table}[!htbp]
    \centering
    \caption{Annotations of Driving Environment Datasets}
    \resizebox{\linewidth}{!}{
    \begin{tabular}{ccccc} 
        \toprule
        Dataset &\makecell[c]{Bounding \\ box in FPV} &\makecell[c]{Bird-view \\Physical States} &\makecell[c]{HD \\ map} &\makecell[c]{Behavior \\ Labels} \\
        \midrule
        \textcolor{black}{MONA \cite{gressenbuch2022mona}}  &\textcolor{black}{$\times$} &\textcolor{black}{\checkmark} &\textcolor{black}{\checkmark} &\textcolor{black}{$\times$} \\
        \textcolor{black}{nuPlan \cite{Caesar2021nuplan}}  &\textcolor{black}{\checkmark} &\textcolor{black}{\checkmark} &\textcolor{black}{\checkmark} &\textcolor{black}{$\times$} \\
        \textcolor{black}{DARI V2X-seq \cite{yu2023v2x}}  &\textcolor{black}{\checkmark} &\textcolor{black}{\checkmark} &\textcolor{black}{\checkmark} &\textcolor{black}{$\times$}\\
        ApolloScape \cite{huang2018apolloscape}   &\checkmark &$\times$ &$\times$ &$\times$ \\
        nuScenes \cite{caesar2020nuscenes}  &\checkmark &$\checkmark$ &\checkmark &$\times$ \\
        WAYMO\cite{sun2020scalability}  &\checkmark &\checkmark &\checkmark &\checkmark \\
        WoodScape\cite{yogamani2019woodscape}  &\checkmark &\checkmark &\checkmark &$\times$ \\
        KITTI \cite{geiger2012we} &\checkmark &$\times$ &\checkmark &$\times$ \\
        Lyft L5 \cite{houston2020one}  &$\times$ &\checkmark &\checkmark &$\times$ \\
        NGSIM \cite{kovvali2007video} &$\times$ &\checkmark &$\times$ &$\times$ \\
        HighD\cite{krajewski2018highd}  &$\times$ &$\times$ &$\times$ &\checkmark \\
        ROAD \cite{singh2021road} &\checkmark &$\times$ &$\times$ &\checkmark \\
        Argoverse \cite{chang2019argoverse}  &\checkmark &\checkmark &\checkmark &$\times$ \\
        PIE\cite{rasouli2019pie}  &\checkmark &$\times$ &$\times$ &\checkmark \\
        TITAN \cite{malla2020titan}  &\checkmark &$\times$ &$\times$ &\checkmark \\
        DrivingStereo\cite{yang2019drivingstereo}   &$\times$ &$\times$ &\checkmark &$\times$ \\
        ONCE\cite{mao2021one}  &\checkmark &$\times$ &\checkmark &$\times$ \\
        H3D\cite{patil2019h3d}  &\checkmark &\checkmark &\checkmark &$\times$  \\
        HDD \cite{ramanishka2018toward}  &\checkmark &$\times$ &$\times$ &\checkmark  \\
        CADC\cite{pitropov2021canadian}  &\checkmark &\checkmark &$\times$ &$\times$ \\
        BDD100K \cite{yu2020bdd100k} &\checkmark &$\times$ &\checkmark &$\times$  \\
        INTERACTION \cite{zhan2019interaction} &\checkmark &\checkmark &\checkmark &$\times$ \\
        PREVENTION \cite{izquierdo2019prevention} &\checkmark &$\times$ &$\times$ &\checkmark \\
        DDD20\cite{hu2020ddd20} &\checkmark &$\times$ &$\times$ &\checkmark \\
        A*3D \cite{pham20203d} &\checkmark &$\times$ &\checkmark &$\times$ \\
        \bottomrule
    \end{tabular}}
    \label{Table:Environment Datasets Annotations}
\end{table}

As a summary, the sensor selection is diverse among different datasets currently, which makes it difficult to develop generalized decision-making methods across different datasets. Besides, as for annotations, the labels of behaviors are not unified.

\subsection{Data Scenarios}
NGISM and KITTI are two of the earliest naturalistic  driving datasets related to driving environment description. They collected vehicle trajectories either from FPV or BV \cite{kovvali2007video} \cite{geiger2012we}. The large volumes of data have promoted the development of driving assistance system and \textcolor{black}{autonomous vehicles} greatly. However, one disadvantage is that in these two datasets, the majority of recorded scenarios are regular ones such as car-following and lane-changing, failing to test the system’s ability to drive in dangerous occasions \cite{krajewski2018highd}. Therefore, in recent years more and more researchers have been focusing on complicated driving environments.

\begin{figure}[!t]
\centering
\includegraphics[width=\linewidth]{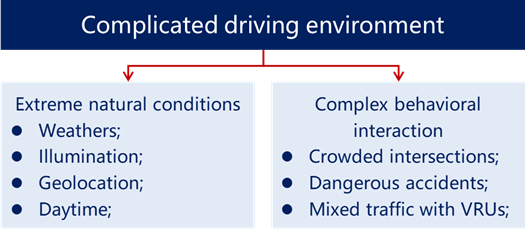}
\caption{Two kinds of complicated driving environment}
\label{Fig:Driving Environment}
\end{figure}

As demonstrated in Fig. \ref{Fig:Driving Environment}, extreme conditions can be divided into two aspects. One is about natural conditions, and the other is about complex behavioral interactions. Ligocki et al. \cite{ligocki2020brno} used infrared cameras to shoot videos in extreme conditions such as dark night and heavy rain. Yang et al. \cite{yang2019drivingstereo} proposed the DrivingStereo dataset that involved nearly all kinds of weather and illumination conditions, like cloudy and ducky days. Boreas is a multi-season \textcolor{black}{autonomous driving} dataset featured by involving all four seasons and many extreme kinds of weather \cite{burnett2022boreas}. In general, data in challenging nature conditions are becoming more and more mature.

\begin{table}[!htbp]
    \centering
    \caption{\textcolor{black}{Datasets including complex behavioral interactions}}
    \resizebox{\linewidth}{!}{
    \begin{tabular}{cc} 
        \toprule
        Dataset &Behavioral Interaction Descriptions\\
        \midrule
        ApolloScape \cite{huang2018apolloscape} &\makecell[c]{Three level of complexity divided by the \\number of moving objects in views} \\ \cline{1-2}
        WAYMO \cite{sun2020scalability}  &\makecell[c]{More than 77.5\% scenarios contain \\conflict areas among various agents} \\ \cline{1-2}
        1001 hours \cite{houston2020one} &\makecell[c]{Collect data in crowded places like \\roudabouts, large intersections, etc.} \\ \cline{1-2}
        Argoverse \cite{chang2019argoverse} &\makecell[c]{300 interesting trajectories which satisfy \\at least one preset condition} \\ \cline{1-2}
        PIE \cite{rasouli2019pie}  &\makecell[c]{A interaction possibility to measure the \\complexity level of scenarios} \\ \cline{1-2}
        H3D \cite{patil2019h3d} &\makecell[c]{Challenging scenes with multiple agents \\and partial occlusion}  \\ \cline{1-2}
        HDD \cite{ramanishka2018toward} &\makecell[c]{Scenes within a motion set, including \\turns, merges, etc.}  \\ \cline{1-2}
        INTERACTION\cite{zhan2019interaction}  &\makecell[c]{Data collected in crowded locations such \\as roundabouts and large intersections} \\ \cline{1-2}
        \textcolor{black}{DDD20} \cite{hu2020ddd20} &\makecell[c]{Interactions between vehicle and large \\numbers of pedestrians in public areas} \\ \cline{1-2}
        DoTA \cite{yao2020and} &\makecell[c]{Dangerous scenarios where collisions \\almost happen} \\ \cline{1-2}
        PSI \cite{chen2021psi}  &\makecell[c]{Drivers’ post explanations for specific \\driving behaviors} \\
        \bottomrule
    \end{tabular}}
    \label{Table:Behavioral Interaction}
\end{table}

Another aspect of the complicated environment is behavioral interaction. As we know, most existing AVs did not achieve high-level automation yet \cite{montanaro2019towards}, and one of the critical problems is that decisions in complex interactive traffic scenes are not efficient enough compared to human drivers. Therefore, some studies have been dedicated to developing highly interactive datasets to support the decision-making development aimed at complicated traffic conditions. Waymo extracts complex scenes, defined as those at least involving two conflicting agents, from videos of real-world driving and further \textcolor{black}{divides} them into nine subclasses such as merging, unprotected turns, and intersections \cite{sun2020scalability}. In their dataset, 77.5\% scenarios contain more than two interactive agents. Instead of classifying by agent numbers, the Argoverse dataset collects 300 interesting scenes including intersections, lane-changing, etc. \cite{chang2019argoverse}. From the perspective of traffic safety, the dataset INTERACTION \cite{zhan2019interaction} \textcolor{black}{includes} dangerous interactions in real world, offering materials for learning of decision making under extreme scenes. Kotseruba et al. \cite{kotseruba2016joint} proposed the JAAD dataset which uses a first-person view to capture the interactions between ego-vehicle and pedestrians. Other datasets that include complex behavioral interactions are summarized in Table \ref{Table:Behavioral Interaction}.

The analysis above shows that complicated traffic interactions have been broadly explored in recent large-scale datasets. However, unlike the simple classification of natural conditions, complex behavior interaction is still vague to describe. As illustrated above, the definitions and features of interaction among various datasets differ from each other. The standards for judging whether interaction exists are also different, e.g., route conflictions \cite{sun2020scalability} \cite{kotseruba2016joint}, specific sets of actions [39], the possibility of hazard \cite{zhan2019interaction}, or just presence of surroundings agents \cite{patil2019h3d}. Guo et al. proposed the concept of drivability consisting of scene drivability, a binary collision indicator, and discretized behaviors to indicate the interaction complexity \cite{guo2019safe}. But the evaluation method \textcolor{black}{is} still coarse and incomplete. Therefore, how to design ideal datasets to make them really interactive is still a topic worth discussing.

\subsection{Data Labeling Techiniques}
Unlike the data related to vehicles or drivers, the source data of traffic environment including raw videos, LiDAR point clouds, etc., usually cannot be used by decision-making models directly. How to make correct and reasonable labels of a large amount of data efficiently is challenging.

The majority of annotation methods still rely on human works\cite{guo2019safe} \cite{yin2017use}. However, some auto-labeling models are also developed to accelerate the process. In the ROAD dataset \cite{singh2021road}, a 3D-RetinaNet architecture was proposed for online data processing. Houston et al. \cite{houston2020one} developed a labeling toolkit to process sensor data into semantic bird-view trajectories automatically. Latest deep learning methods such as the teacher-student model could realize massive processing on the basis of a small amount of manually labeled data frames \cite{tarvainen2017mean}.

\begin{figure}[!t]
\centering
\includegraphics[width=\linewidth]{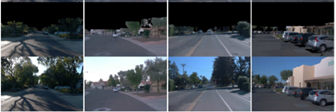}
\caption{Top row: Synthesized realistic data; Bottom row: Real camera image}
\label{Fig:Data Labeling}
\end{figure}

A novel approach to \textcolor{black}{labeling} data is synthesized realistic technology \cite{yang2020surfelgan}. Data sensors in the real world have blind zones of perception, resulting in some data defects. To overcome the difficulty, Yang et al. presented an effective method to generate a digital twin scenario based only on limited amounts of LiDAR and camera on real vehicles so that developers could extract desired target data at each spot and time of the traffic event , as shown in Fig. \ref{Fig:Data Labeling}. To realize the mapping from real to digit, a deep learning network named SurfelGAN is leveraged to reconstruct the missing states in real the real world.

In conclusion, the labeling technique gradually becomes more and more automatic because of emerging deep-learning-based toolkits. Nevertheless, most behavior labels still rely on manual annotations.

\section{Driver Datasets} \label{Sec:Driver Datasets}
\begin{figure}[htbp]
\centering
\includegraphics[width=\linewidth]{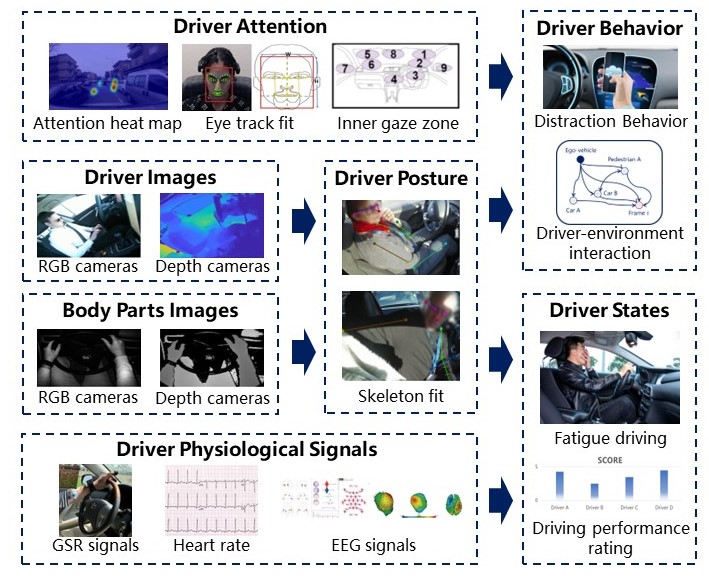}
\caption{An overview of driver datasets}
\label{Fig:driver datasets}
\end{figure}

As autonomous vehicles develop to a higher level, many studies found that learning from human driver intelligence or \textcolor{black}{behaviors} is helpful for better decision-making \cite{ghosh2021speak2label}\cite{xie2021human}. As one source of model input, driver information adds interpretability to the deep-learning-based methods by combining driver logic frameworks into algorithms and can improve decision performance \cite{wang2021towards}\cite{chen2019autonomous}. Hence, datasets concentrated on drivers have gained increasing focus in the last decade. Fig. \ref{Fig:driver datasets}  shows an overview of driver datasets. Data can be generally classified into three categories: driver attention, driver posture, and driver physiological signals. The former two types of data could be used to understand driver behavior including distraction mode \cite{kopuklu2021driver}, interaction \cite{lateef2021saliency}, etc. These labels may act as variables to select corresponding decision models. Driver physiological signals as well as posture contributes to classification of driver states such as fatigue driving \cite{jegham2020novel} and driving performance rating \cite{ling2021driver} \cite{xing2017study}. In this section, driver-related datasets are reviewed in the sequence of attention, posture, and physiological signals.

\subsection{Driver Attention}
Datasets of driver attention mainly focus on the place or points that drivers pay attention to while driving. One  way to collect drivers’ attention data is to use an eye-tracking device to get the area where the drivers look at in the form of gaze maps. A dataset including 555000 frames collected by 8 drivers called Dr(eye)ve \cite{alletto2016dr}, was built in 2016 and used the SMI eye tracking glasses to get gaze maps. CoCAtt \cite{shen2021cocatt}, a dataset built in 2021, used eye-tracking devices to get the difference in driver attention distribution between manual driving and automated driving. Some other ways can also be used to collect drivers’ attention data. Speak2Label \cite{ghosh2021speak2label} applies speak-to-text conversion to automatically label the image-based gaze behavior dataset. Francisco Vicente \cite{vicente2015driver} also utilized a low-cost camera (a Logitech c920 Webcam) placed on top of the steering wheel to get the driver’s images. Then together with Parameterized Appearance Models and a system named SDM, images can be transferred to detect and track drivers’ facial features, as well as estimate drivers’ head pose.

\begin{figure}[htbp]
    \begin{minipage}[t]{0.45\linewidth}
        \centering
        \includegraphics[width=\textwidth]{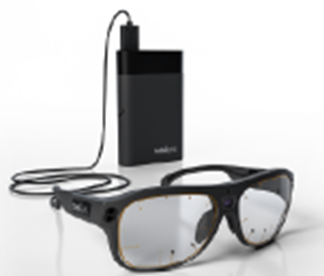}
        \centerline{(a)}
    \end{minipage}%
    \begin{minipage}[t]{0.55\linewidth}
        \centering
        \includegraphics[width=\textwidth]{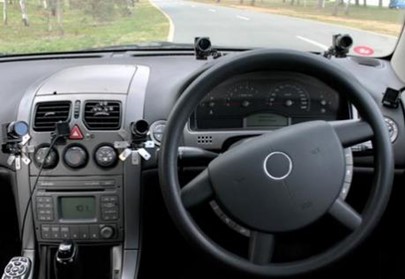}
        \centerline{(b)}
    \end{minipage}
    \caption{(a) is the glasses-formed device, and (b) is the fixed form}
    \label{Fig:driver devices}
\end{figure}

Two kinds of eye-tracking devices are commonly used to collect attention data, as shown in Fig. \ref{Fig:driver devices}. Glasses-formed eye-tracking device offers a first-person view. Cameras integrated into the glasses capture video the driver sees, and it rotates synchronously with the driver’s head movement, ensuring sufficient view cover. One disadvantage is that the glasses hinder drivers from totally nature driving, especially for those who have worn their own glasses. Another type uses multiple cameras installed above the dashboard  of vehicles. By perception from multi-cameras, eye movements can be extracted. With external installed devices, the drivers will not be disturbed. However, the fixed-cameras form is not capable of tracking eye movement when the driver looks backward or sideways.

\subsection{Driver Posture}

Driver posture datasets are summarized in Table \ref{Table:Driver Datasets}. The data of driver images record the video of driver’s body or driver’s hands, which can be used to judge whether the driver is focusing on the road and can take over the control of the vehicle or not. Cameras fixed in cabins are widely-used tools to collect data. For example, the datasets DriPE \cite{guesdon2021dripe} and Driver Anomaly Detection(DAD) \cite{kopuklu2021driver} use RGB cameras to record driver behaviors. Apart from RGB cameras, depth cameras could enrich datasets and increase the accuracy of driver pose detection. Drive\&Act \cite{martin2019drive} applies multi-modal sensors to collect various types of driver behavior data, such as color, infrared, depth and 3D body pose information. 3MDAD \cite{jegham2020novel} derive data from two Microsoft Kinect cameras, both including color (RGB) sensors, infrared (IR) sensors, and microphones, and recordes data with an acquisition rate of 30fps. Focusing on the behaviors of drivers’ hands, Nikhil Das \cite{das2015performance} built a video-based driver hands dataset with different background complexities, illumination settings, users, and viewpoints. The dataset includes over 2000 annotated images from three testing vehicles and over 2000 images from YouTube which were recorded in unknown vehicles.

\begin{table}[!htbp]
    \centering
    \caption{Data Types of Driving Posture Datasets}
    \resizebox{\linewidth}{!}{
    \begin{tabular}{cccc} 
        \toprule
        Dataset &\makecell[c]{Body Parts \\ Images} &\makecell[c]{Driver \\ Images} &\makecell[c]{Driver \\ Posture} \\
        \midrule
        Speak2Label \cite{ghosh2021speak2label}  &$\times$ &\checkmark &$\times$ \\ 
        Brain4Cars \cite{jain2016brain4cars}  &$\times$ &\checkmark &$\times$ \\
        Drive\&Act \cite{martin2019drive}  &\checkmark &\checkmark &\checkmark \\
        \makecell[c]{Driver Anomaly \\Detection (DAD) \cite{kopuklu2021driver}} &\checkmark &\checkmark &$\times$ \\
        3MDAD \cite{jegham2020novel}  &$\times$ &\checkmark &$\times$ \\
        VIVA \cite{das2015performance} &$\times$ &\checkmark &$\times$ \\
        DriPE \cite{guesdon2021dripe}  &$\times$ &\checkmark &\checkmark \\
        \bottomrule
    \end{tabular}}
    \label{Table:Driver Datasets}
\end{table}

The labels of driver posture is important to driver-related datasets.  DriPE \cite{guesdon2021dripe} annotates driver postures manually by a bounding box and 17 key points including arms and legs each with 3 key points and 5 additional makers for the eyes, ears, and nose. The dataset Drive\&Act \cite{martin2019drive} obtains 3D body pose via triangulation of 2D poses from 3 frontal views. The 3D head poses are obtained by employing the OpenFace algorithm. The hand detection dataset VIVA uses the Aggregate Channel Features(ACF) object detector to extract hand pose.

\subsection{Driver Physiological signals}
This category of driver data is deeply connected with traffic psychology. Physiological signals are commonly used to judge the driver’s mental state and the level of various emotions such as nervousness, anxiety, fatigue, etc. \textcolor{black}{Some research \cite{fleicher2022using} \cite{ayoub2022real} \cite{dillen2020keep} shows that driver or passenger physiological signals could be an input to improve the performance of decision-making algorithms.} According to the physiological indicators, the driver’s performance can also be evaluated so that the distinguished ones are filtered out.

Ling et al \cite{ling2021driver} utilized electrocardiogram (ECG) and galvanic skin responses (GSR) to collect heart rates and skin temperatures. With these reflective inputs, they assessed driver cognitive loads by feature analysis. Xing et al \cite{xing2017study} established a human-in-the-loop driving simulator based on the Ergolab system, a comprehensive platform that collects skin, muscle, breath, and blood oxygen signals. Remier et al \cite{reimer2014effects} added vocal signal into the driver database by installing sensors on the front window. Besides, in Foy’s research, the participants of experiments were asked to wear an \textcolor{black}{Electroencephalogram (EEG)} to collect the activation levels of prefrontal cortex.

The amount of driver physiology research is sufficient, however, no large-scale public dataset is available for human driver intelligence extraction. The main challenge is that few datasets provide time-aligned driving data of the vehicle, so no ground truth can be used as the training target of \textcolor{black}{decision-making} methods.

\subsection{Limitation of Driver Datasets}
Datasets for driver analysis mentioned above have two main shortages. The sizes of these datasets are relatively small, and the usage of driver datasets is also limited.

The sizes of the driver-related datasets are not large generally, and the scenes contained in these datasets are not abundant. For example, datasets about driver attention are mainly collected in urban scenes, which have clear traffic signs and good traffic conditions. Datasets about driver poses usually consider the influence of lighting conditions instead of traffic conditions, which also limits the usage of these datasets. Another limitation of current driver datasets is the incompleteness of data interfaces. None of the existing datasets covered all driver attention, posture, and physiological information simultaneously, which restricted potential applications on decision-making.

In conclusion, current driver-related datasets are still limited both in data volume and sources\textcolor{black}{. And} current datasets mainly focus on one single driver data type, which can’t be used in the research in which driver action, attention and physiological information are all needed.

\section{Dataset Applications} \label{Sec:Dataset Applications}

\begin{figure}[!t]
\centering
\includegraphics[width=\linewidth]{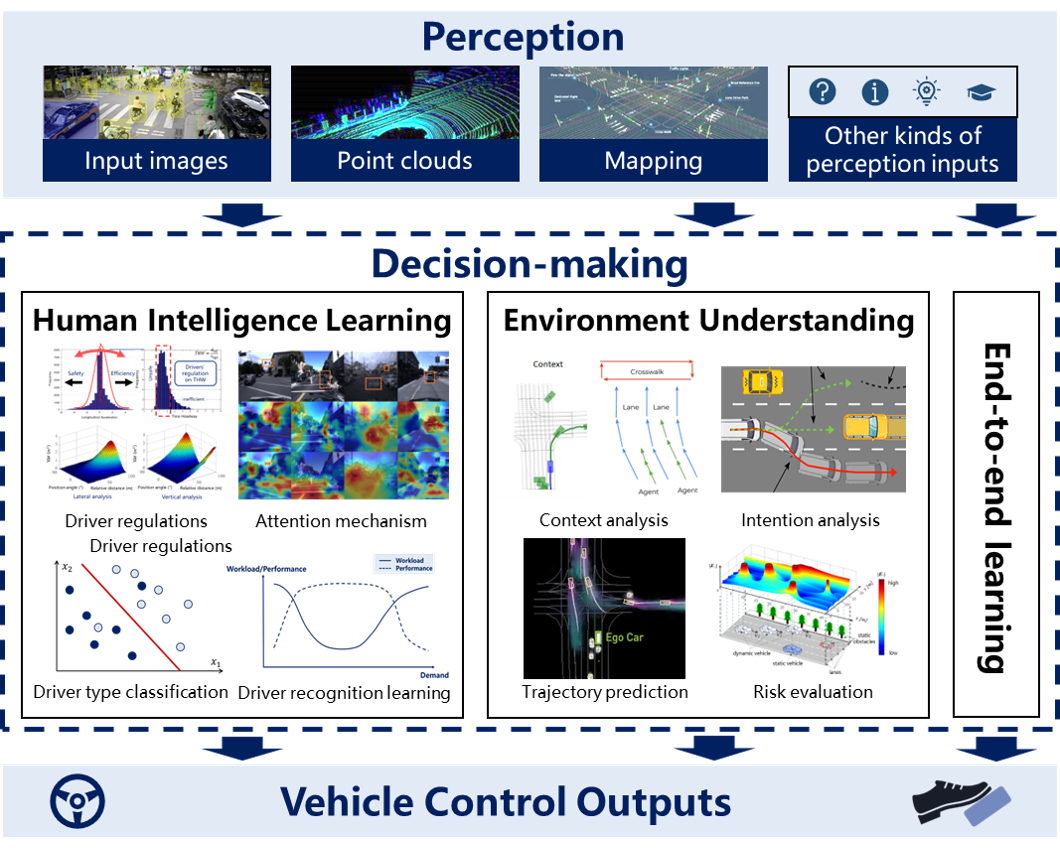}
\caption{Different technical routes of decision-making}
\label{Fig:decision routes}
\end{figure}

With the development of decision-making technology, the data types useful for decision-making algorithms are also expanded. The technical route of research on data-driven decision-making methods varies greatly, and the routes are summarized in Fig. \ref{Fig:decision routes}. Some research tries to learn from human intelligence \cite{xie2020develop}, some research directly uses the raw perception data to obtain decisions via end-to-end algorithms \cite{hecker2018end}, while some research applies semantic environment understanding to guarantee better interpretability \cite{oh2017object}. Each technical route has its own unique advantages and shortcomings, and also require different datasets. Therefore, choosing proper datasets for different research is a fundamental step but may also be confusing for beginners.  Considering the previous survey research did not pay much attention to the applications of datasets for decision-making, this section illustrates the potential applications of the mentioned datasets for decision-making.

Based on the different technical routes shown in Fig. \ref{Fig:decision routes}, the applications of the introduced datasets for decision-making can be briefly divided into three aspects: \textcolor{black}{datasets for driver intelligence learning and analysis}, environment understanding and decision learning. There are several subdivisions and typical datasets of each aspect as Table \ref{Table:Dataset Applications} shows, which are detailed in the following.

\begin{table}[!htbp]
\renewcommand\arraystretch{1.2}
    \centering
    \caption{\textcolor{black}{Dataset Applications, Subdivisions, and Typical Datasets}}
    \resizebox{\linewidth}{!}{
    \begin{tabular}{ccc} 
        \toprule
        Applications &Subdivisions &Typical Datasets\\
        \midrule
        \multirow{2}{*}{\makecell[c]{\textcolor{black}{Driver Intelligence}\\\textcolor{black}{Learning and Analysis}}}  &Driver Attention &Dr(eye)ve \cite{alletto2016dr}, CoCAtt \cite{shen2021cocatt}, Speak2Label \cite{ghosh2021speak2label} \\  \cline{2-3}
        &Driver Behaviors &DriPE\cite{guesdon2021dripe}, DAD  \cite{kopuklu2021driver}, Drive\&Act \cite{martin2019drive}, VIVA \cite{das2015performance}\\ \cline{1-3}
        \multirow{3}{*}{\makecell[c]{Environment \\Understanding}}  &Context Analysis &highD \cite{krajewski2018highd}, DoTA\cite{yao2020and}, BDD100K \cite{yu2020bdd100k}, RUGD \cite{wigness2019rugd} \\ \cline{2-3}
        &Intention Analysis &PIE \cite{rasouli2019pie}, PSI \cite{chen2021psi}, JAAD \cite{kotseruba2016joint}, HDD \cite{ramanishka2018toward}\\ \cline{2-3}
        &Trajectory Prediction &Argoverse (2) \cite{chang2019argoverse} \cite{wilson2021argoverse}, PREVENTION \cite{izquierdo2019prevention}\\ \cline{1-3}
        \multirow{4}{*}{\makecell[c]{Decision \\Learning}}  &Behavioral Strategy &D$^2$CAV \cite{toghi2020maneuver}, HDD \cite{ramanishka2018toward}, ROAD \cite{singh2021road}, Brain4Cars \cite{jain2016brain4cars} \\ \cline{2-3}
        &Trajectory Planning &\makecell[c]{NGSIM \cite{kovvali2007video}, KITTI \cite{geiger2012we}, DrivingStereo \cite{yang2019drivingstereo}, \\WAYMO \cite{sun2020scalability}, INTERACTION \cite{zhan2019interaction}}\\ \cline{2-3}
        &Manipulation Learning &DBNet \cite{chen2019dbnet}, DDD17 \cite{binas2017ddd17}, DDD20 \cite{hu2020ddd20}, A2D2 \cite{geyer2020a2d2}\\
        \bottomrule
    \end{tabular}}
    \label{Table:Dataset Applications}
\end{table}

\subsection{\textcolor{black}{Datasets for Driver Intelligence Learning and Analysis}}
\textcolor{black}{Driver intelligence learning and analysis} based on relative datasets could help autonomous vehicles to learn human-like decision-making. \textcolor{black}{Datasets for driver intelligence learning and analysis are mainly distinguished by their adequate data of drivers, and} can be roughly divided into three classes: driver attention, driver behavior, and driver operation. Datasets for driver attention mainly focus on the places or points to which drivers pay more attention while driving. Datasets for driver behavior mainly collect the drivers’ actions and movement data, which could be used to detect and predict the status of the driver. As the third class, driver operation includes the basic driver inputs and their results, changing and overtaking, the corresponding dataset will be introduced in the part of the \textcolor{black}{manipulation learning} dataset in Section \ref{Subsubsec:manipulation learning}. In the following we focus on the former two applications.

\subsubsection{Driver Attention}
Drivers usually pay more attention to places where dangerous situations might happen. By analyzing drivers’ attention mechanism, the autonomous vehicle system could focus on the riskier areas for better safety given limited perception and computing resource, or at least could detect the distraction situation of drivers and provide warning. 

For example, the dataset Dr(eye)ve \cite{alletto2016dr} can be applied to understand how the driver’s past experience may influence driver’s attention on different objects in the scene. The data can also be used to judge whether drivers keep their attention on the traffic or not while the autonomous driving system is working. Another dataset CoCAtt \cite{shen2021cocatt} can also be used to study the relationship between driver attention and autonomous driving system [66]. The speak-to-text conversion applied in the dataset Speak2Label \cite{ghosh2021speak2label} could estimate the drivers’ gaze zone roughly. 

In general, datasets for driver attention mostly collect the attention data of drivers, and can be utilized by research on the analysis of driver’s attention area and distraction problem, to help the autonomous driving system to recognize important areas and be more focused.

\subsubsection{Driver Behaviors}
Considering the fact that the current autonomous driving technique is not mature enough to cover all the situations, which means drivers still need to keep their eyes on the road to be ready to take over the vehicle at any time, driver behavior datasets can help to identify the driver’s statues, reflect driver's decision, assess the complication level of the scenarios the driver is faced up with, judge whether the driver should take over the vehicle when the autonomous driving system fails to work well, etc. For manual-driving vehicles, it can also help estimate the posture, tiredness, and attention status of drivers.

The annotation of the dataset DriPE\cite{guesdon2021dripe}, including a bounding box and 17 key points, could benefit the research of driver posture estimation. Furthermore, some datasets focus on anomaly situations in favor of the practicability of the algorithms. The Driver Anomaly Detection (DAD) dataset  \cite{kopuklu2021driver} distinguishes anomalous driving situations to help researchers test the performance of the driver behavior detection algorithm under abnormal conditions. The multi-modal data collected in the Drive\&Act dataset \cite{martin2019drive} enables driver behavior research by different sensors.  Drivers’ hands are usually essential in driving. The VIVA dataset contains annotated images of the drivers’ hands for hand detection, provides another way to estimate drivers’ status.

In summary, there is a variety of valuable driver behaviors that are necessary to be detected to improve the security of both manual driving and autonomous driving. Related datasets and research are also vivid and provide various data and methods to train brilliant models. 

\subsection{Datasets for Environment Understanding}
Environment understanding is an indispensable link between environment perception and motion planning. For autonomous vehicles, the environmental information collected by the perception module is usually processed by the environment understanding module, to identify drivable areas, detect the intention of pedestrians, predict the future trajectory of road users, and so on. 

\textcolor{black}{In this subsection, relative datasets contain not only the perception data, but also results after environment understanding, which is categorized into three main subdivisions: context analysis, intention analysis, and trajectory prediction.}

\subsubsection{Context Analysis}
We apply the word “context” here to indicate the factors or participants in the driving scenario related to the ego vehicle \cite{tivesten2015driving}. This topic includes research on scenario classification, anomaly detection, semantic segmentation, drivable area detection, and so on.

The highD dataset \cite{krajewski2018highd} assesses the context by criterion such as Time to Collision (TTC) and Time Headway (THW), which can be utilized for a system-level validation of highly automated driving systems. The DoTA dataset \cite{yao2020and} contains 18 anomaly categories of driving video and multiple anomaly participants in different driving scenarios, which is significantly helpful for learning a robust detector to analyze anomaly context. Some other datasets can be applied to analyze road information and semantic segmentation. The BDD100K dataset \cite{yu2020bdd100k} containing 100K videos and 10 context analysis tasks can be used for lane marking detection, and furthermore distinguishing drivable areas for autonomous vehicles. Besides, the dataset also benefits heterogeneous multitask learning, including lane marking, semantic instance segmentation, and so on.  For driving on unstructured roads, identifying drivable areas is difficult and suffers from a lack of semantically labeled datasets compared to their urban counterparts. The RUGD dataset \cite{wigness2019rugd} which contains different and complicated areas of unstructured driving scenarios such as dirt, sand, grass, and trees, can help solve this unique challenge. 

\subsubsection{Intention Analysis}
The intention of traffic participants is another significant criterion in decision-making. A number of datasets are designed to identify or predict the intention of pedestrians and vehicles.

Pedestrian behavior or intention anticipation is a key challenge in the design of assistive and autonomous driving systems. The PIE dataset \cite{rasouli2019pie} contains large-scale videos and models for pedestrian intention estimation and trajectory prediction. Specifically, this dataset concentrates on pedestrians that tend to interact with ego-vehicle and the behavior or intention is annotated by human experts. With the development of natural language processing techniques, the text-based explanation of driver’s understanding of the scenes and the reasoning process are recorded in the PSI dataset \cite{chen2021psi}. The explanations can link the critical visual scene features with the human decision-making process and results, enabling the development of explainable and human-like pedestrian intention analysis and prediction algorithms. Instead of focusing on pedestrians only, the JAAD dataset \cite{kotseruba2016joint} is produced for the analysis on all surrounding road users. The HDD dataset that has 4-layer annotation structure\cite{ramanishka2018toward} can help the exploration of interactions between drivers and traffic participants from cause and effect labels. 

\subsubsection{Trajectory Prediction}
The trajectory prediction of other transportation participants is also an essential task for autonomous driving, for which plenty of datasets have arisen.

The Argoverse dataset \cite{chang2019argoverse} can be used for 3D tracking and motion forecasting of vehicles that are in unusual scenarios such as at intersections, taking left or right turns, and changing to adjacent lanes in dense traffic. Moreover, the Argoverse 2 Dataset \cite{wilson2021argoverse} expands the categories of objects to 30 from 17 of the Argoverse 1, objects such as strollers and wheelchairs are included to provide richer samples for trajectory prediction algorithm design. The dataset PREVENTION \cite{izquierdo2019prevention} uses multi-modality of sensors, which ensures both redundancy and complementarity of information and also labels all the actions that appear in the scene. Such a dataset can be used to accelerate research on trajectory prediction under strong interaction.

\subsection{Datasets for Decision Learning}
\begin{figure}[!t]
\centering
\includegraphics[width=\linewidth]{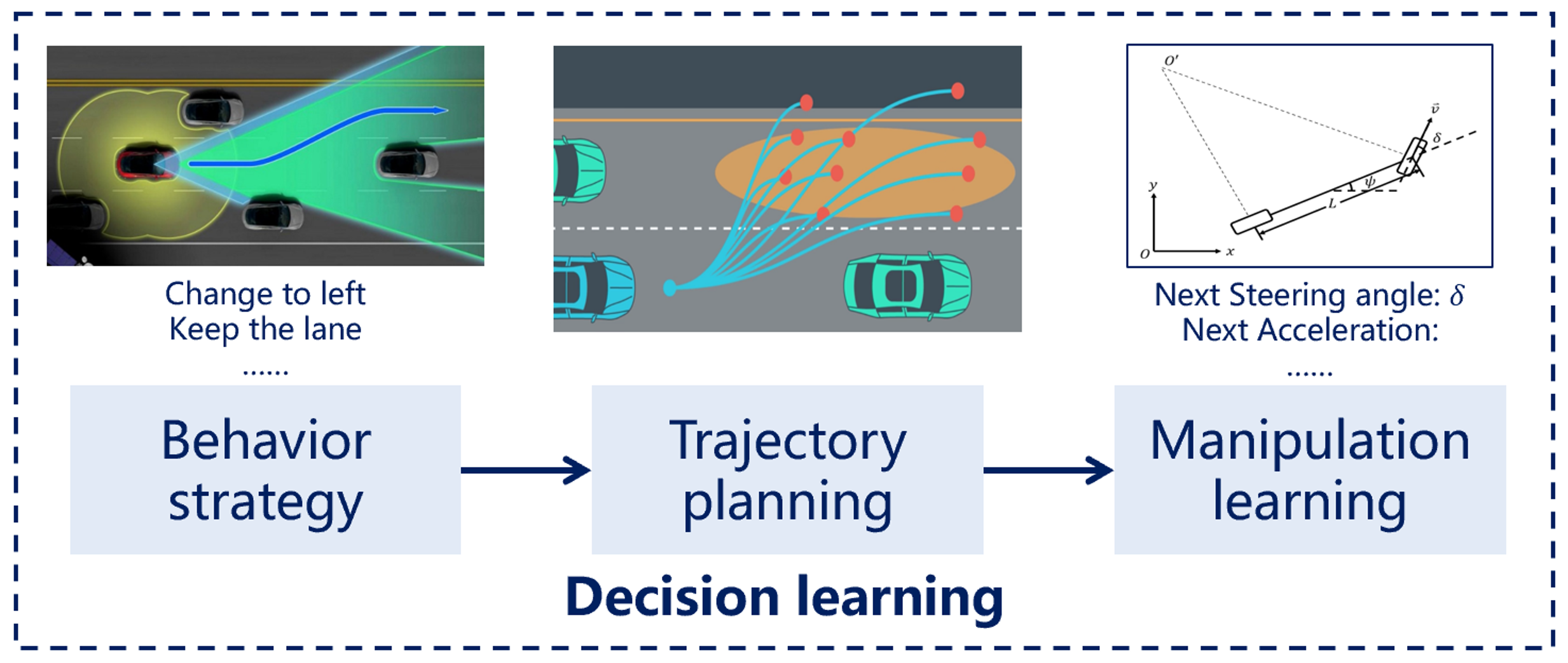}
\caption{\textcolor{black}{Subdivisions of decision learning levels}}
\label{Fig:decision learning}
\end{figure}

As powerful deep learning models spring up, learning-based methods show their promising potential to cope with complicated driving contexts and scenarios \cite{huang2020autonomous}. An efficient and appropriate dataset is a must for decision learning research. Decision learning can be divided into three submodules: behavioral strategy, trajectory planning, and \textcolor{black}{manipulation learning} \cite{furda2011enabling}, as shown in Fig. \ref{Fig:decision learning}. \textcolor{black}{Datasets that can be applied to these three submodules are usually featured by the decision level data, and they are analyzed, compared, and concluded in the following.}

\subsubsection{Behavioral Strategy}
As for data-driven algorithms, behavioral strategy learning is usually realized by classification machine learning methods. These methods provide discrete behaviors instead of continuous motions, thus learning autonomous driving decisions is based on the behavior annotation of vehicles in the datasets. 

The D$^2$CAV dataset \cite{toghi2020maneuver} labels events into discrete behaviors including turns, lane shifts, deceleration, etc. The ratios of different behaviors are balanced in D$^2$CAV to mimic reality. Comparatively, the 4-layer annotation structure in HDD dataset \cite{ramanishka2018toward} and the triplet structure of road event in the ROAD dataset \cite{singh2021road} provide sufficient antecedents and consequences to better express a traffic event, making it easier for the model to learn precise behavioral strategies and possess interpretability.  As for modality, the Brain4Cars dataset \cite{jain2016brain4cars} captures drivers’ conditions together with road environment information, by which joint effect models could be developed to make behavior decisions.

To learn practical behavioral strategy, it is necessary for datasets to reflect the real situations in data collection. Besides, detailed annotation and processing of raw data could help algorithms to improve interpretability.

\subsubsection{Trajectory planning}
Trajectory planning is an essential step for autonomous driving. In related datasets, actual trajectories of ego vehicle could be directly used as the ground truth of model training processes.

The two most classical datasets used for trajectory planning model training is NGSIM \cite{kovvali2007video} and KITTI \cite{geiger2012we}. However, most scenarios in these two datasets are regular and safe, failing to test the system’s ability to drive on dangerous occasions. Therefore, in recent years, more and more research has been focusing on challenging and extreme conditions. The diverse scenarios and weathers contained in the DrivingStereo dataset \cite{yang2019drivingstereo} equips the trajectory planning research with robustness in extreme conditions. Another important characteristic of the complicated environment is interaction. The Waymo dataset \cite{sun2020scalability} and the INTERACTION dataset \cite{zhan2019interaction} both provide abundant interactive scenarios to benefit relative research.

Datasets for trajectory planning are one of the most adequate \textcolor{black}{autonomous driving} datasets for decision-making. The scene richness and interactivity are becoming promising in dataset collection and algorithm learning.

\subsubsection{Manipulation learning} \label{Subsubsec:manipulation learning}
The third level of decision learning is manipulation learning. \textcolor{black}{As for current vehicle structures, the final control input are the wheel steering angle, throttle position, and brake position. To learn these manipulation level decision strategies, onboard information needs to be recorded.}

A standard data structure is like the DBNet dataset \cite{chen2019dbnet}, which contains both onboard and environmental information for manipulation learning. Besides standard data, other internal data is also collected in related datasets to assist the algorithm training. DDD17 \cite{binas2017ddd17} and DDD20 \cite{hu2020ddd20} offer more complete onboard data, such as engine speed, odometer, etc. Similarly, A2D2 \cite{geyer2020a2d2} records all data on the CAN bus involving even internal status like liquid pressure of brake valves. 
The data structure of datasets for \textcolor{black}{manipulation learning} is rather simple. Extra types of data in addition to the fixed structure may help the algorithm learning procedure.

\section{Future Trends} \label{Sec:Future Trends}
It is obvious that datasets about drivers, environments, and vehicles are vital for developing decision-making systems of high-level autonomous vehicles. Many organizations have been devoted to collecting comprehensive data from various aspects. In this section, we summarize three future trends for potential improvements.

\begin{enumerate}
    \item Collecting data from all the driver, environment, and the ego vehicle simultaneously. As aforementioned, human intelligence is an important reference for high-level autonomous vehicles \cite{xie2021human} \cite{chen2019autonomous}. Models like Brain4Cars \cite{jain2016brain4cars} have proved that drivers’ attention and facial information helped to make better decisions. However, as introduced above, currently the human-factor-related data is isolated from those of vehicles and the environment. Without alignments on timelines and locations, it is difficult to use driver information as inputs for elaborate decision methods. Therefore, a dataset containing all driver, environment, and vehicle information can complement the vacancy, exploring more possibilities for AV decision-making by extraction of  human’s driving intelligence.
    \item Containing data from more sensor sources. Most datasets are developed for certain tasks. To solve the target problem in the most efficient way, the main concerns about the selection of sensors are costs and operational complexity. As a result, the data included usually just covers the requirements for the developers' research interest. This is suitable for specific fields of research, but not friendly for generalized applications. Third-party users of datasets demand various data such as Radar, event cameras, infrared, etc, especially more and more end-to-end methods have emerged. Consequently, a comprehensive and ideal dataset should contain as many commonly used sensors as possible to offer data redundancy and generalization. \textcolor{black}{Furthermore, as the rapid development of electric vehicles, it will be promising to include the battery SOC (state of charge) and electromotor status in the autonomous vehicle decision-making datasets to promote research about energy-saving decision-making of autonomous vehicles.}
    \item Establishing a unified and compatible data format. At present, the data formats of datasets are different. Some datasets provide processed information in json, csv, txt, etc., some offer data in a self-defined struct, and others upload the raw sensor sources with annotations attached. As the complexity and size of parameter spaces of decision-making models grow larger and larger, the required amount of data also explodes to a new level. An ideal pattern is that the training and modeling process can be finished across various datasets with no or little specific adaption. However, data utilization across different datasets could not be realized at this stage because of the heterogeneous data formats. Accordingly, datasets with unified  standard and interfaces will greatly promote the development of intelligent vehicles.
\end{enumerate}

\section{Conclusion} \label{Sec:Conclusion}
In this paper, we survey the state-of-the-art decision-making datasets of autonomous vehicles from four aspects: vehicle-related data, environment-related data, driver-related data, and their specific applications. After briefly reviewing the developing history of decision datasets, we present a detailed summary including many latest ones which have not been covered in previous surveys. We classify the data types, annotations, and features into specific categories. Based on comparison and feature analysis, limitations and potential applications on various decision-making of autonomous vehicles are concluded. By organizing the review in the categories of driver, vehicle, and environment datasets, this survey is dedicated to provide a clear and intuitive selection reference for researchers and developers of AV decision-making when the algorithm developing requires data. Also future trends are summarized to give advice on future dataset collection forms.

\bibliographystyle{IEEEtran}
\bibliography{IEEEabrv, mylib}


\begin{IEEEbiography}
[{\includegraphics[width=1in,height=1.25in,clip,keepaspectratio]{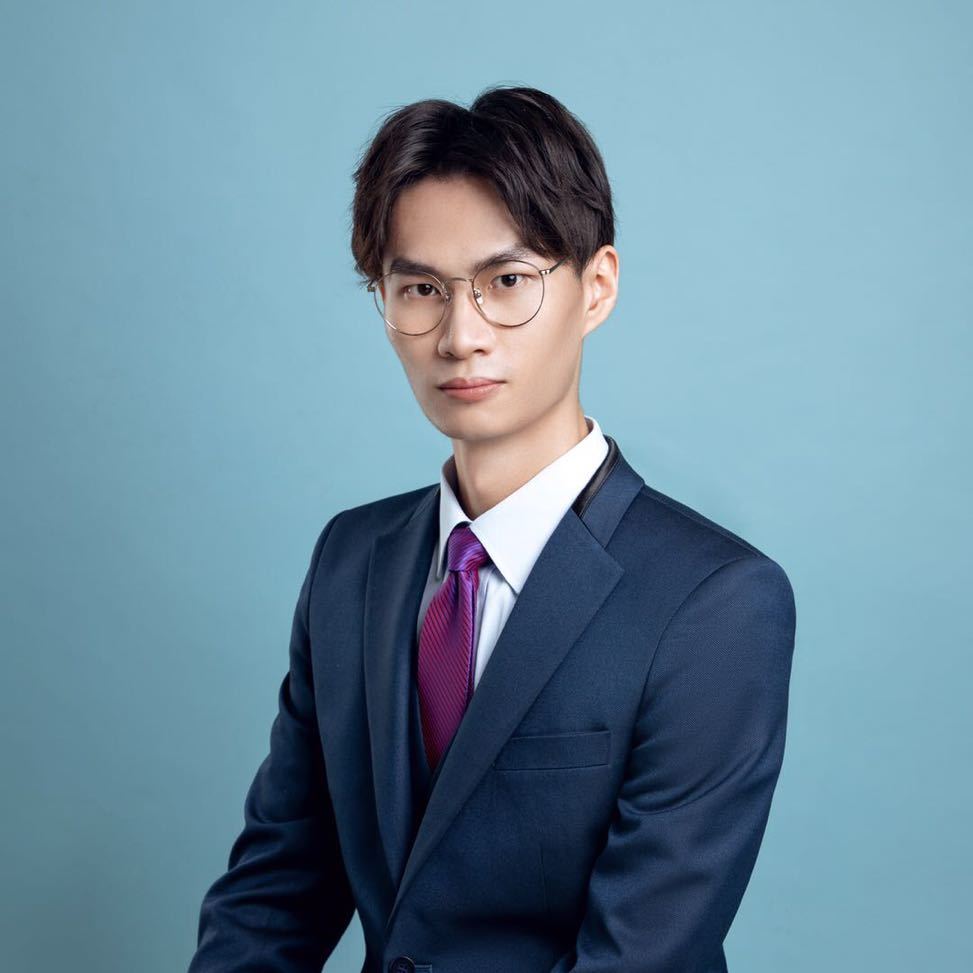}}]
{Yuning Wang} 
received the bachelor’s degree in automotive engineering from School of Vehicle and Mobility, Tsinghua University, Beijing, China, in 2020. He is currently pursuing the Ph.D. degree in mechanical engineering with School of Vehicle and Mobility, Tsinghua University, Beijing, China. His research centered on risk evaluation, trajectory prediction and decision-making process of intelligent and connected vehicles.
\end{IEEEbiography}

\begin{IEEEbiography}
[{\includegraphics[width=1in,height=1.25in,clip,keepaspectratio]{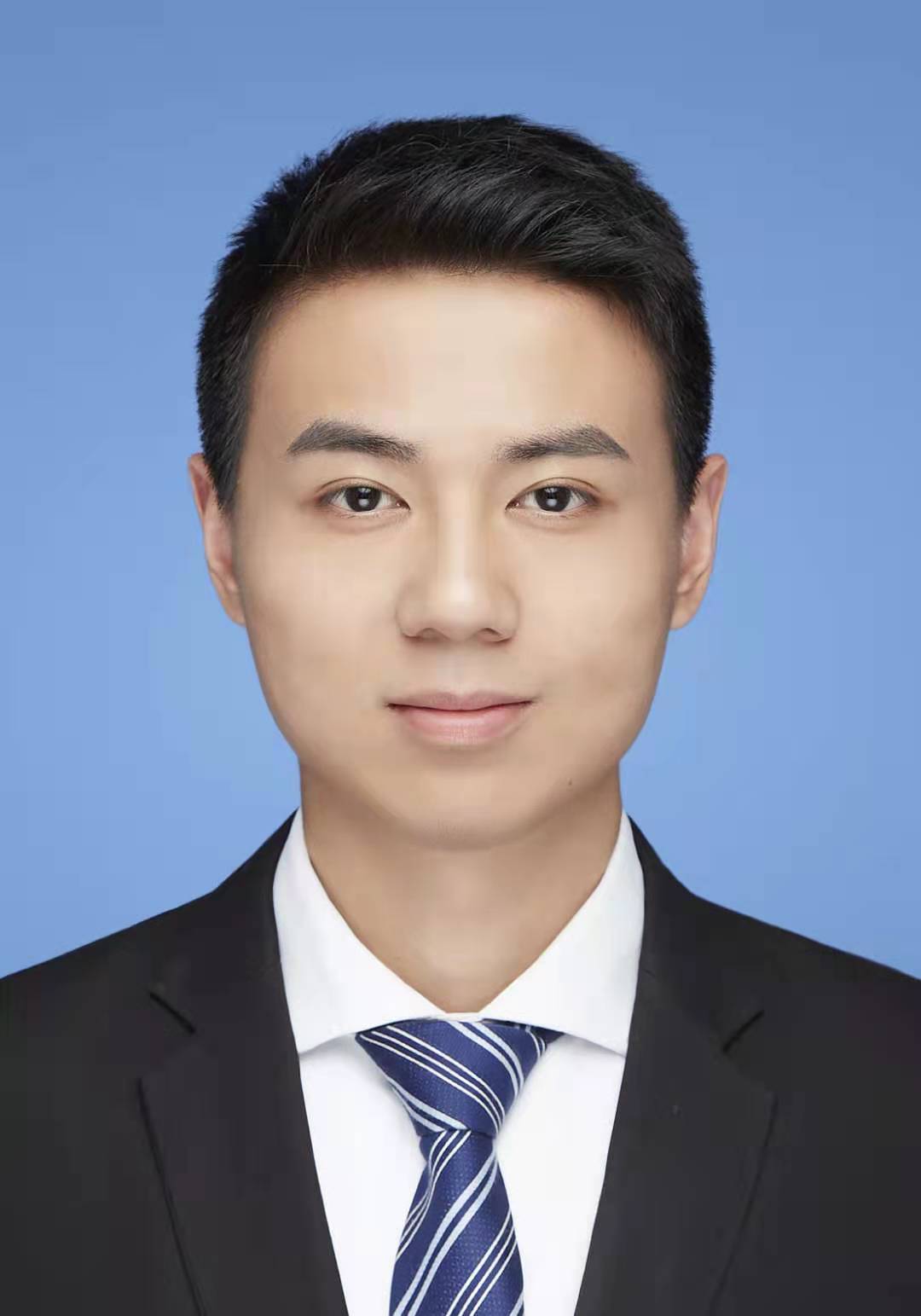}}]
{Zeyu Han} 
received the bachelor’s degree in automotive engineering from School of Vehicle and Mobility, Tsinghua University, Beijing, China, in 2021. He is currently pursuing the Ph.D. degree in mechanical engineering with School of Vehicle and Mobility, Tsinghua University, Beijing, China. His research interests includes risk evaluation, environment understanding of intelligent vehicles.
\end{IEEEbiography}

\begin{IEEEbiography}
[{\includegraphics[width=1in,height=1.25in,clip,keepaspectratio]{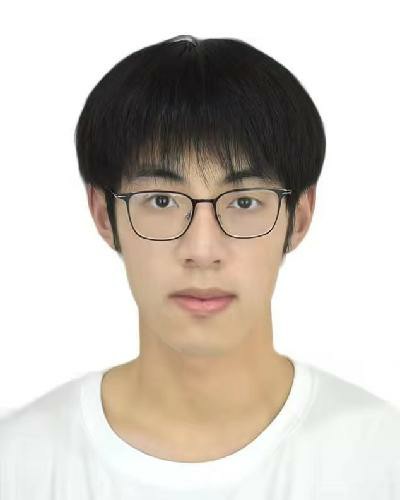}}]
{Yining Xing} 
received the bachelor’s degree in automotive engineering from School of Vehicle and Mobility, Tsinghua University, Beijing, China, in 2022. He is currently pursuing the Ph.D. degree in mechanical engineering with School of Vehicle and Mobility, Tsinghua University, Beijing, China. His research centered on risk evaluation, trajectory prediction and decision-making process of intelligent vehicles.
\end{IEEEbiography}

\begin{IEEEbiography}
[{\includegraphics[width=1in,height=1.25in,clip,keepaspectratio]{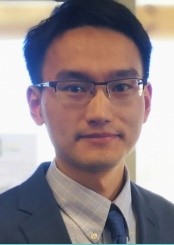}}]
{Shaobing Xu} 
received his Ph.D. degree in Mechanical Engineering from Tsinghua University, Beijing, China, in 2016. He is currently an assistant professor with the School of Vehicle and Mobility at Tsinghua University, Beijing, China. He was an assistant research scientist and postdoctoral researcher with the Department of Mechanical Engineering and Mcity at the University of Michigan, Ann Arbor. His research focuses on vehicle motion control, decision making, and path planning for autonomous vehicles. He was a recipient of the outstanding Ph.D. dissertation award of Tsinghua University and the Best Paper Award of AVEC’2018.
\end{IEEEbiography}

\begin{IEEEbiography}
[{\includegraphics[width=1in,height=1.25in,clip,keepaspectratio]{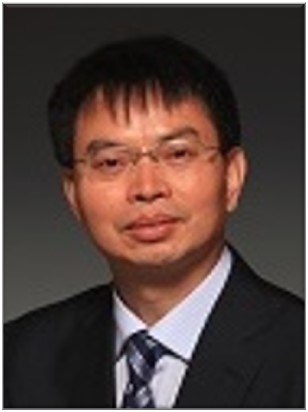}}]
{Jianqiang Wang} 
received the B. Tech. and M.S. degrees from Jilin University of Technology, Changchun, China, in 1994 and 1997, respectively, and Ph.D. degree from Jilin University, Changchun, in 2002. He is currently a Professor of School of Vehicle and Mobility, Tsinghua University, Beijing, China. He has authored over 150 papers and is a co-inventor of over 140 patent applications. He was involved in over 10 sponsored projects. His active research interests include intelligent vehicles, driving assistance systems, and driver behavior. He was a recipient of the Best Paper Award in the 2014 IEEE Intelligent Vehicle Symposium, the Best Paper Award in the 14th ITS Asia Pacific Forum, the Best Paper Award in 2017 IEEE Intelligent Vehicle Symposium, the Changjiang Scholar Program Professor in 2017, Distinguished Young Scientists of NSF China in 2016, and New Century Excellent Talents in 2008.
\end{IEEEbiography}

\vfill

\end{document}